\documentclass[letterpaper, 10 pt, conference]{ieeeconf}  

\IEEEoverridecommandlockouts                              

\usepackage[table,dvipsnames]{xcolor}
\usepackage{svg}
\usepackage{color}
\newcommand{\red}[1]{\textcolor{red}{#1}}

\usepackage{amssymb}
\usepackage{graphicx}
\usepackage{subcaption}
\usepackage[group-separator={\,}]{siunitx}
\usepackage{booktabs} 
\usepackage{amsmath}

\usepackage{amsthm}
\usepackage{bm}
\usepackage{multirow}
\let\labelindent\relax
\usepackage{enumitem}
\usepackage{multicol}
\usepackage{hyperref}
\usepackage{nicefrac}
\usepackage{tabularx}
\usepackage{soul}
\sethlcolor{ForestGreen!25}

\usepackage{pgfplots,lipsum}
\usepackage{scalefnt}
\usepackage{pgfplotstable}
\pgfplotsset{compat=newest}

\title{\LARGE \bf
Agent-Agnostic Centralized Training\\ for Decentralized Multi-Agent Cooperative Driving

}

\author{Shengchao Yan$^{1}$, Lukas König$^{2}$ and Wolfram Burgard$^{3}$
\thanks{The authors are with the $^1$Department of Computer Science, University of Freiburg, Germany, the $^2$Institute for Neural Computation, Ruhr University Bochum, Germany, and the $^3$Department of Engineering, University of Technology Nuremberg, Germany.}%
}

\begin{document}

\maketitle
\thispagestyle{empty}
\pagestyle{empty}

\begin{abstract}

Active traffic management with autonomous vehicles offers the potential for reduced congestion and improved traffic flow. However, developing effective algorithms for real-world scenarios requires overcoming challenges related to infinite-horizon traffic flow and partial observability. To address these issues and further decentralize traffic management, we propose an asymmetric actor-critic model that learns decentralized cooperative driving policies for autonomous vehicles using single-agent reinforcement learning. By employing attention neural networks with masking, our approach efficiently manages real-world traffic dynamics and partial observability, eliminating the need for predefined agents or agent-specific experience buffers in multi-agent reinforcement learning. Extensive evaluations across various traffic scenarios demonstrate our method's significant potential in improving traffic flow at critical bottleneck points. Moreover, we address the challenges posed by conservative autonomous vehicle driving behaviors that adhere strictly to traffic rules, showing that our cooperative policy effectively alleviates potential slowdowns without compromising safety.

\end{abstract}

\section{Introduction}
Traffic congestion is a prevalent issue in various parts of our road system, such as intersections, ramps, and lane drops, and significantly undermines traffic efficiency. It increases accident risks, fuel consumption, emissions, and higher driver frustration and discomfort~\cite{ha2020leveraging}. To alleviate congestion, research has extended beyond designing better road infrastructure to include innovative traffic management strategies, from implementing variable speed limits~\cite{alessandri1999nonlinear} to enhancing traffic signal intelligence~\cite{yan2020efficiency}, utilizing road sensors to feed information to centralized units which, in turn, issue directives to drivers. However, the deployment of such centralized control systems is often hindered by the costs and complexity associated with their construction and maintenance.

The advent of autonomous vehicles (AVs) presents a promising shift away from traditional traffic management approaches and towards introducing more efficient methods that capitalize on the capabilities of AVs for perception, communication, and decision-making~\cite{ma2016freeway,yan2021courteous,wu2022flow}. This evolution suggests the possibility of substituting road sensors with the distributed sensing and communication capabilities of AVs, and replacing dynamic traffic signals with direct commands to AVs, thereby simplifying the interaction with human-driven vehicles (HVs) which continue to follow standard traffic rules.

Despite their effectiveness in simulations, centralized traffic management methods face significant challenges in real-world applications. Issues such as limited bandwidth for communication between AVs and control units and susceptibility to adverse weather conditions can undermine their effectiveness. To address these challenges and obviate the need for centralized control, recent research has explored decentralized decision-making based on AVs' local observations, with multi-agent reinforcement learning (MARL) emerging as a popular approach for decentralized vehicle control~\cite{ha2020leveraging,vinitsky2023optimizing,schester2021automated,dai2023socially}. This approach has demonstrated success across various domains, including gaming~\cite{vinyals2019grandmaster}, traffic light control~\cite{prabuchandran2014multi}, and resource scheduling optimization~\cite{xu2020multi}. However, applying MARL directly to traffic management is challenging, especially in accurately representing the infinite-horizon and dynamic nature of traffic flow.

This paper introduces an asymmetric actor-critic model~\cite{pinto2018asymmetric} to learn decentralized cooperative driving policies through single-agent reinforcement learning.
By integrating attention neural networks~\cite{vaswani2017attention} with masking, the novel actor-critic architectures can adeptly manage variable traffic inputs and partial observability.
It is worth noting that while our method utilizes single-agent algorithms, it also fits within the centralized-training-decentralized-execution framework \cite{lowe2017multi} of MARL.
Unlike conventional MARL algorithms, our approach is agent-agnostic, as it treats each AV as a token of the system state, rather than as distinct agents.
This eliminates the need for a predefined set of agents or the maintenance of agent-specific experience buffers, a challenge in traffic environments where the agent number can grow indefinitely with the episode length.

We test our approach rigorously against conventional controllers in realistic traffic scenarios across various road system features, including intersections, ramps, and lane drops. The findings highlight the capacity of our method to substantially enhance traffic flow using decentralized policies and partial observations. Additionally, we investigate the impact of conservative AV driving behaviors~\cite{zhan2016non} and demonstrate how our cooperative policy can effectively mitigate these concerns, paving the way for a safer, more efficient, and adaptable traffic management paradigm.

\section{Background and Related Work}
\label{sec:related_work}

\subsection{Single-Agent and Multi-Agent Reinforcement Learning}
    Reinforcement Learning (RL) enables an agent to learn decision-making by interacting with its environment, modeled as a Markov Decision Process (MDP).
    At each step with state $s\in\mathcal{S}$, the agent selects an action $a\in\mathcal{A}$ according to the observation $o\in\mathcal{O}$ and policy $\pi(\cdot\mid o)$.
    The system then transitions to a new state $s'\in\mathcal{S}$ according to the transition probability $P(s'\mid s,a)$ and receives a scalar reward $r\in\mathbb{R}$. Overall, the agent aims at maximizing the expected discounted cumulative reward $\max_{\pi} \mathbb{E}_{\pi, P} \left[ \sum_{t=0}^{+\infty} \gamma^t R(s_t, a_t) \right]$.

    Multi-agent reinforcement learning (MARL) extends RL for environments with multiple interacting agents, represented by Markov games. A Markov game is a tuple $\left\langle \mathcal{N}, \mathcal{S}, \mathcal{O}, \mathcal{A}, P, R_{i}, \gamma \right\rangle$, where $\mathcal{N}$ is the set of all agents,  $\mathcal{O}_i$ and $\mathcal{A}_i$ are observation space and action space for agent $i$, and $\mathcal{O} = {\times}_{i \in \mathcal{N}} \mathcal{O}_{i}$ and $\mathcal{A} = {\times}_{i \in \mathcal{N}} \mathcal{A}_{i}$ represents the joint observation and action space. 
    Each agent $i$ maintains an individual policy and reward function.
    Let $\Pi_i = \left\{ \pi_i(a_i\mid o_i): \mathcal{O}_i \to \Delta_{\mathcal{A}_i} \right\}$ be the policy space for agent $i$, then the objective for agent $i$ is represented as $\max_{\pi_i} \mathbb{E}_{\pi, P} \left[ \sum_{t=0}^{+\infty} \gamma^t R_i(s_t, a_t) \right]$.

\subsection{Traffic Management with Reinforcement Learning}
    After the DARPA autonomous vehicle challenges~\cite{behringer2004darpa,buehler2009darpa}, much effort has been taken to develop algorithms for automated driving. This development process was substantially accelerated through the utilization of deep learning approaches.
    Reinforcement learning is mainly adapted by two groups of tasks for vehicle decision making: 1) social navigation~\cite{dai2023socially,zhenghao2021learning,huegle2019dynamic,vinitsky2022nocturne} to learn to navigate through traffic by anticipating the motion of ambient objects; 2) traffic management~\cite{wu2022flow,vinitsky2023optimizing} to improve traffic flow by cooperating with and influencing the behavior of vehicles in the vicinity.
    Although both tasks focus on developing vehicle control policies, they exhibit significant differences. While the navigation goal of the individual AVs is to efficiently reach their target locations, traffic management systems typically aim at an improved overall traffic flow to benefit all participants. Furthermore, traffic management environments generally operate under an infinite horizon, with new vehicles continuously entering the system, while navigation tasks often terminate once the predefined vehicles reach their destinations. Last but not least, to focus on improving system efficiency, traffic management tasks always assume an accident-free environment enabled by collision-checking low-level controllers.

    A substantial amount of work has been published in traffic management systems. Early pioneering work~\cite{wu2017emergent,vinitsky2018benchmarks} utilizes reinforcement learning based on closed-loop maps. Despite the infinite-horizon traffic flow, the considered environments are restricted due to their fixed set of vehicles. Moreover, these works also assume full observability of and perfect communication between the AVs.
    Others adopt MARL to account for partial observability and a variable number of agents~\cite{ha2020leveraging,vinitsky2023optimizing}. However, they are restricted to a predefined set of agents~\cite{vinitsky2023optimizing}.
    Although the idea to reroute the released AVs back to the map entrance makes it possible for infinite-horizon traffic input, the flow rate is not able to vary due to the fixed number of AVs. Moreover, agents could exploit the unrealistic model by learning to predict the reappearance of other AVs in the scenario.
    In this paper, we aim to tackle a broad spectrum of challenges in traffic management, including partial observability, infinite-horizon traffic dynamics, and a fluctuating number of vehicles.

\subsection{Safety and Cautiousness in Autonomous Driving}
    Critical traffic flow bottlenecks require careful interactions between AVs and other road users to ensure safety.
    Strictly following traffic rules with excessively cautious behavior, however, may lead to inefficiencies and increased wait times~\cite{CNN2021Waymo}. 
    This is why several papers raise the question of whether AVs should sometimes trade off safety for efficiency similar to human drivers~\cite{zhan2016non,basu2017do,leurent2020safe}. 
    This complex issue has yet to be thoroughly explored for infinite-horizon traffic flow.
    We propose to use a decentralized policy to mitigate the drawbacks of conservative AV behaviors through collaboration, without compromising safety standards.

\section{Methods}
\label{sec:methods}

\begin{figure}
    \vspace{5pt}
    \centering
    \includegraphics[width=0.4\textwidth]{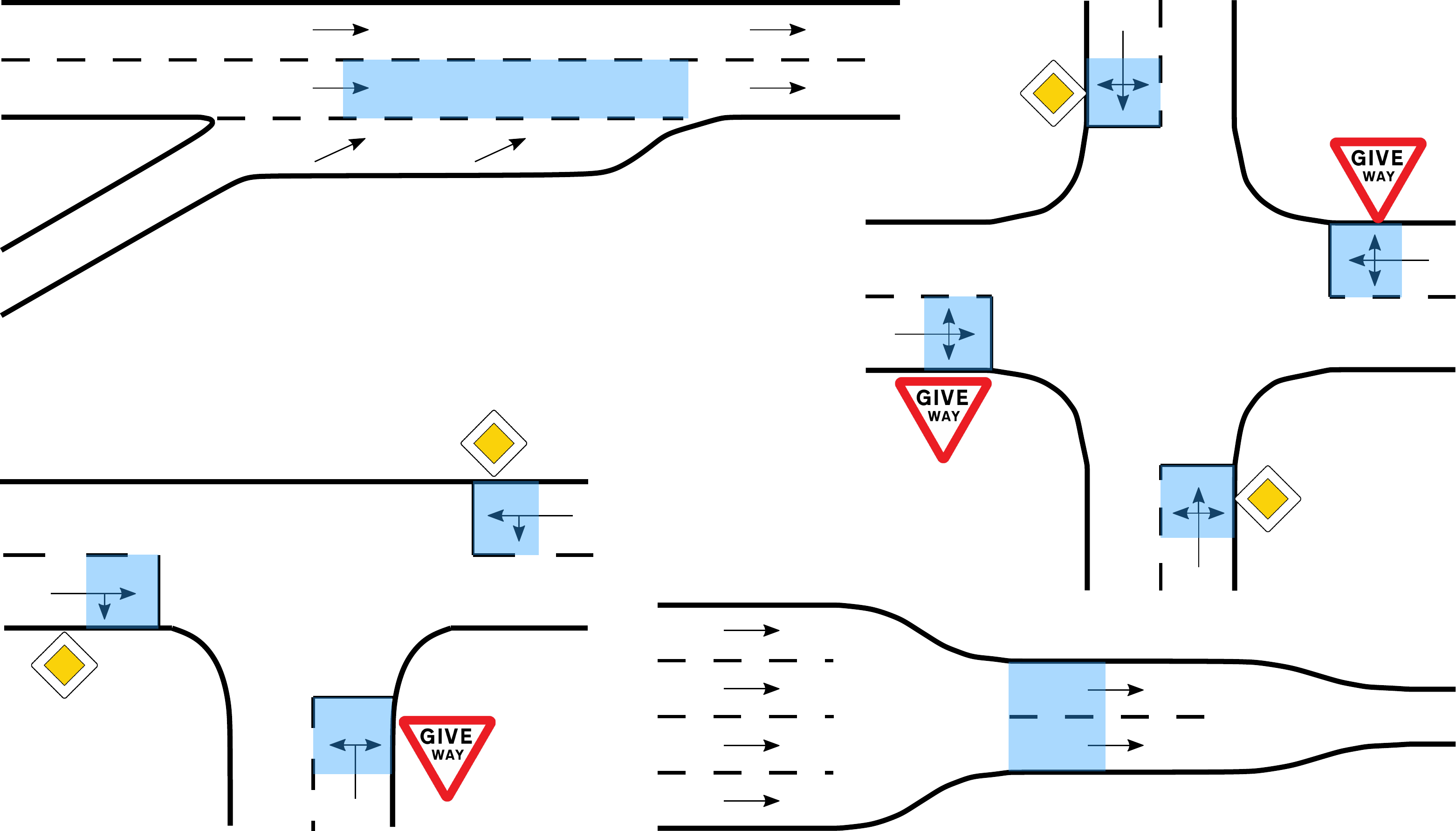}
    \caption[centric]{Common traffic bottlenecks: on-ramp merge, four-way intersection, three-way intersection, lane drop. AVs follow the learned policy only in the blue areas as described in Sec.~\ref{sec:state_space}.}
    \label{fig:maps}
\end{figure}

Our method aims to solve the traffic management problem in different bottleneck scenarios visualized in Fig.~\ref{fig:maps}.
We propose a novel actor-critic model, that uses asymmetric inputs to learn a decentralized cooperative driving policy for individual AVs.
Within this section, we will describe the state and action spaces as well as the reward function and will provide a detailed description of the asymmetric actor-critic that allows for partial observability, infinite-horizon traffic input, and a varying number of vehicles.

\subsection{State, Observation, Action and Reward}
    \begin{figure}
        \vspace{5pt}
        \centering
        \includegraphics[width=0.4\textwidth]{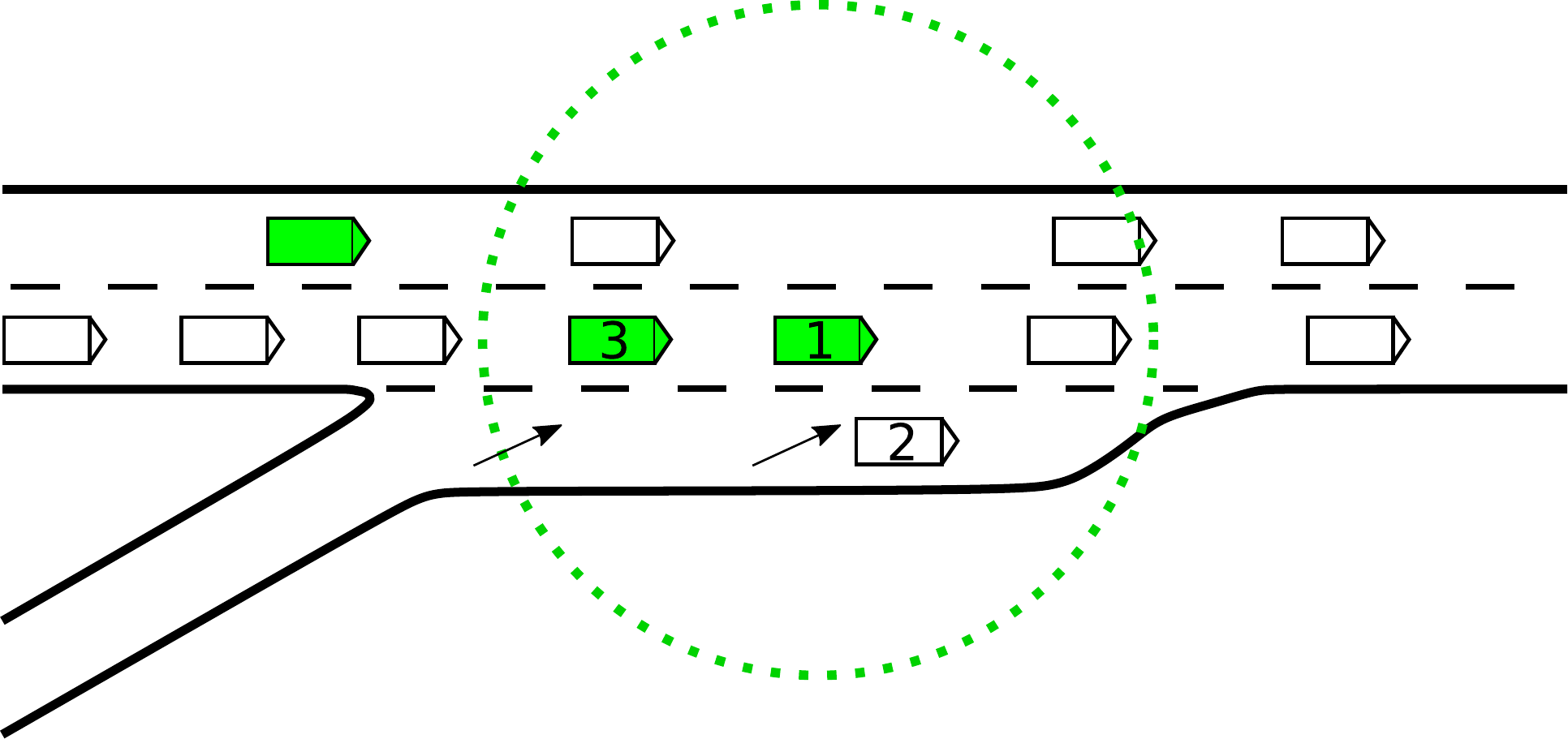}
        \caption[centric]{Vehicle~$2$ intends to merge into a dense freeway. Green vehicles are AVs, while white ones are HVs.
        The dashed circle represents the sensing range of vehicle $1$.
        A gap for vehicle~$2$ to merge in can be created by either lane changing of AV~$1$ or slowing down of AV~$3$.}
        \label{fig:onramp}
        \vspace{-5pt}
    \end{figure}

    We consider the task in standard reinforcement learning settings. Proximal policy optimization (PPO)~\cite{schulman2017proximal} is used as the backbone algorithm. 
    The standalone bottleneck locations shown in Fig.~\ref{fig:maps} are the primary focus of this work. The problem of on-ramp merge is visualized in Fig.~\ref{fig:onramp} as an example.
    
    \subsubsection{State Space}
    \label{sec:state_space}
    The state of the scenario consists of a mask indicating existing vehicles and the features of all vehicles. 
    The state mask $M_s$ is a boolean vector of dimension $C$, where $C$ is the capacity corresponding to the maximum number of vehicles this scenario can hold. Each value indicates the existence of one vehicle.
    The state feature $F_s$ is represented by a 2D vector of dimension $C\times d_\mathrm{v}$, where $d_\mathrm{v}$ is the length of the vehicle feature.
    The feature vector of each vehicle is composed of eight values: $(x,y,\sin(\alpha),\cos(\alpha),v,l, c,t)$. Here, $x$ and $y$ represent the position of the vehicle in the map (normalized by the dimensions of the map), and $\alpha$ represents the angle of its heading direction. The term $v$ stands for the velocity, which is normalized by the speed limit, while $l$ is the status of the turn signal, which uses values $\{-1,0,1\}$ for right-turning, no signal, and left-turning.
    Vehicles' routes are randomly selected at the beginning of each episode.
    The term $c$ represents the vehicle category and can take the values $\{-1,0,1\}$ for HV, inactivated AV, or activated AV. The term $t$ is the travel time of the AV in seconds since it entered the map, which is normalized with an empirical value of $300$. The HV travel time is defined as $-1$ based on the assumption that only AVs can record and communicate this information.
    
    Only AVs near the bottleneck points are regarded as activated and as driving according to the learned policy, since the cooperative behavior mostly happens here. Other AVs follow the default driver models~\cite{behrisch2015sumo,salles2020extending}, which are commonly used in traffic simulators.
    Including all AVs in policy training would flood the training data with information on a single modality. We select the activated AVs empirically on certain lane segments (see Fig.~\ref{fig:maps}) and leave it as future work to automate this process.

    \subsubsection{Observation Space}
    To accommodate the partial observability of the decentralized policy, each AV is limited to acquiring features from nearby vehicles within its sensing range.
    The observation of the scenario is composed of a mask indicating activated AVs and an observation mask.
    The AV mask $M_\mathrm{AV}$ is a boolean vector of dimension $N$, where $N$ is the maximum number of activated AVs.
    The observation mask $M_\mathrm{obs}$ is a 2D boolean vector of dimension $N\times C$, where each row indicates the observed vehicles of each AV.
    Using masks instead of extracting the observed vehicle features can reduce computation and memory load. Combined with the attention-based actor-critic, this observation representation contributes to vectorizing the inference of the reinforcement learning model.
    Similar to real-world traffic conditions, vehicles' destinations are not in the observation and can only be inferred through the use of turn signals.
    
    \subsubsection{Action Space}
    The joint action space $\mathcal{A}$ has a dimension of $N\times d_a$, where $d_a$ is the number of discrete actions of each vehicle.
    A 2D boolean vector of this dimension is given as an action mask $M_a$.
    Although this is a large action space, the policy sharing among AVs, which is explained in Sec.~\ref{sec:actor_critic}, enhances the training process by reducing the exploration space~\cite{yan2023geometric}.
    Additionally, the action mask is utilized to further reduce the exploration difficulty.
    The vehicles in the rightmost lane, for example, do not have the action of changing to the right lane.
    The discrete action space of each vehicle consists of six actions $\{a_\mathrm{left},a_\mathrm{right},a_{v_0},a_{v_1},a_{v_2},a_{v_3}\}$, where $a_\mathrm{left}$ and $a_\mathrm{right}$ stand for changing into the left or the right lane, and $a_{v_i}$ represents adjusting the velocity to $v_i$. In this work, four target velocities $\{0,0.33,0.66,1\}\times v_\mathrm{limit}$ are chosen to give vehicles more flexibility during cooperation while still forcing it to obey the speed limit $v_\mathrm{limit}$.
    We note that the actions only represent high-level driving intentions. An AV selects an action every second and attempts to execute lane-changing within the next $5$ seconds after choosing $a_\mathrm{left}$ or $a_\mathrm{right}$. The intention terminates either upon a successful lane change or when the 5-second period expires.
    The low-level control of the vehicles is handled by the simulator so this method is focused on traffic management in a collision-free environment.
    Combined with formal safety verification, the hierarchical control strategy is beneficial for developing safe and reasonable autonomous driving policies~\cite{mirchevska2018high}.
    Incorporating a broader set of target velocities, acceleration, or even utilizing a continuous space are all viable approaches. Essentially, it represents a trade-off between the flexibility of actions and the simplicity of exploration.

    \subsubsection{Reward Function}
    In our previous research~\cite{yan2021courteous}, we introduced a centralized controller to manage intersections of mixed traffic with AVs and HVs. The controller used a dedicated reward function to balance the interests of individual vehicles against the broader objective of improving overall traffic flow.
    This paper adopts the same throughput-based reward-shaping strategy to consider both fairness (equity) and operational efficiency in traffic management:
    \begin{align}
        r_t = \eta_\text{b} + \eta_\text{a}\cdot\sum_{i=1}^{N^{\text{TP}}_t}\,{\tau_i},
    \end{align}
    where $\eta_a$ and $\eta_b$ are the linear equity factors, $\tau_i$ is the travel time of the $i\mathrm{th}$ released vehicle in time step $t$, and $N^{\text{TP}}_t$ represents the number of released vehicles in time step $t$.

\subsection{Asymmetric Actor Critic}
\label{sec:actor_critic}
    \begin{figure}[t]
        \vspace{5pt}
        \centering
        \includegraphics[width=0.4\textwidth]{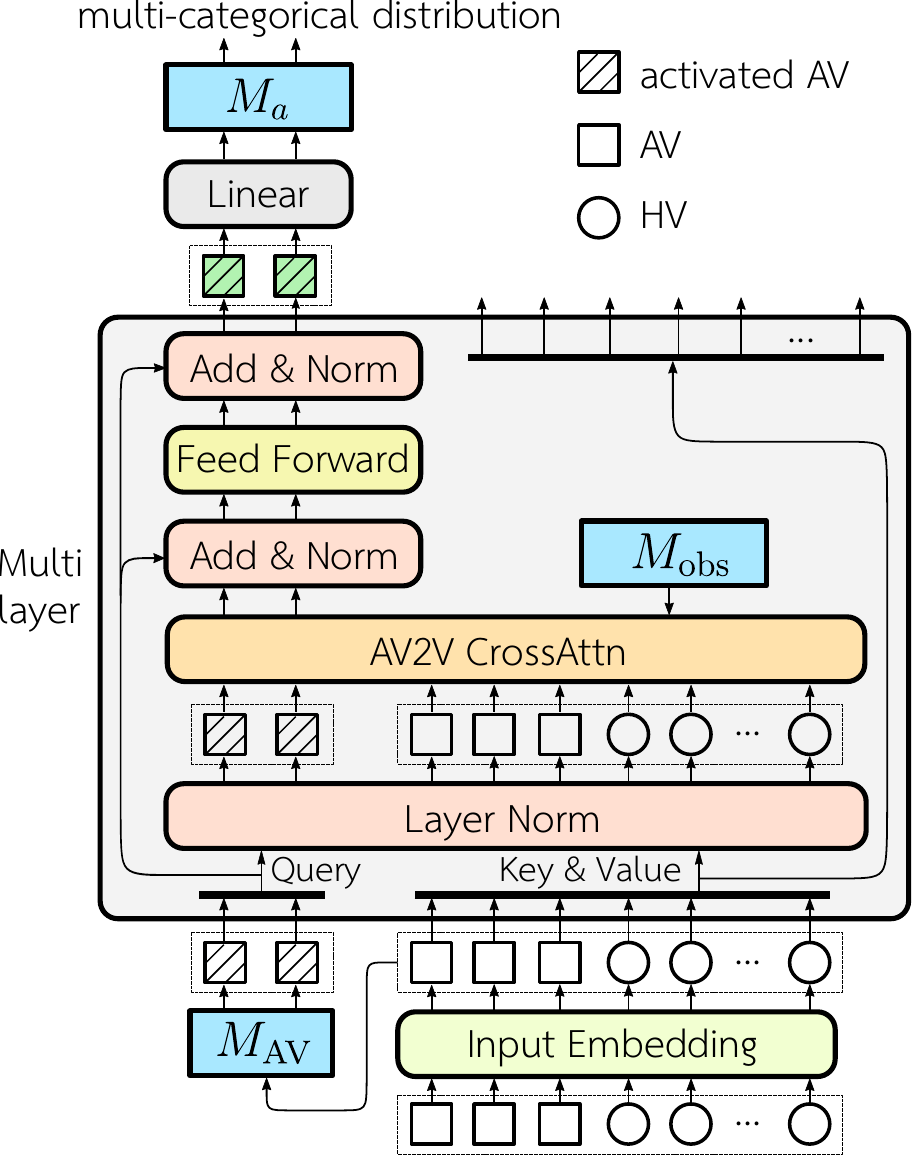}
        \caption[centric]{Policy network. The network input is from the on-ramp scenario visualized in Fig.~\ref{fig:onramp}, where two AVs out of three are activated.}
        \label{fig:policy}
        \vspace{-5pt}
    \end{figure}

    Methods have been proposed to handle a variable number of observed vehicles~\cite{huegle2019dynamic, leurent2019social}.
    To manage the variability in the number of observers (i.e. AVs) within the environment, researchers frequently leverage multi-agent reinforcement learning~(MARL).
    However, existing MARL algorithms possess limitations that render them less effective for traffic management applications, primarily due to
    \begin{itemize}[leftmargin=*]
        \item the restriction to a predefined  set of agents, which does not apply to real traffic with variable vehicle amount growing with episode length and
        \item the requirement that each agent upholds its experience buffer, policy, or critic function, thereby elevating computational and memory demands compared to a streamlined approach utilizing a singular, centralized agent with vectorized calculations.
    \end{itemize}

\begin{figure}[t]
        \vspace{5pt}
        \centering
        \includegraphics[width=0.36\textwidth]{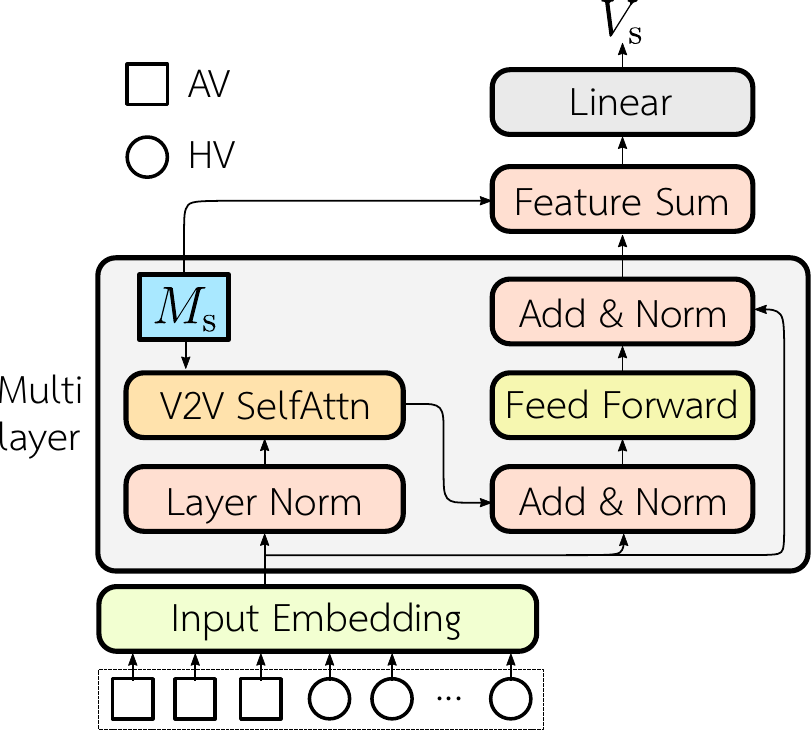}
        \caption[centric]{Critic network. The input embedding layer shares the same parameters with the policy network.
        }
        \label{fig:critic}
        \vspace{-5pt}
    \end{figure}

    In this work, we introduce an asymmetric actor-critic model to meet the previously mentioned traffic management requirements (see Figures~\ref{fig:policy} and~\ref{fig:critic}).
    Although the same state is given to both actor and critic, each AV does not see the whole state. Inputting the state instead of the extracted observation information for each activated AV is beneficial for the vectorized calculation stream. Each activated AV can still only attend to its observed vehicles due to the $M_\mathrm{obs}$ input into the cross-attention layers, making the actor-critic asymmetric.
    By representing each vehicle as an input token of the overall system state rather than an individual agent, as in conventional MARL algorithms, our method can effectively handle infinite-horizon traffic environments with dynamically changing numbers of AVs.

    In the policy network, we first embed the normalized vehicle features with a feed-forward network. Then we select the tokens of the activated AVs with $M_\mathrm{AV}$ and use them as a query for the following attention calculation. We employ the embedded features of all vehicles as key and value.
    The policy network is mainly composed of a stack of two identical attention layers.
    Each layer consists of a cross-attention calculation and a fully connected feed-forward network.
    As suggested by the work on layer normalization~\cite{xiong2020on}, we employ a residual connection after and a layer normalization before each of these sub-layers.
    We only update the tensors of the query in each attention layer, while the key and value always stay the same as the embedded tokens. As a result, AVs do not communicate any information with each other in the policy network, making the policy fully decentralized.
    After the attention layers, we pass the encoded query features through a linear projection to output the action logits for each activated AV. We finally use the logits of dimension $N\times d_a$ to build a multi-categorical distribution $\pi(\bm{a}\mid s)$ for the actions.
    We utilize the mask $M_\mathrm{AV}$ for the calculation of $\pi_\theta(\bm{a_t} \mid s_t)$ and its entropy value in the PPO objective
    to account for only the activated AVs.
    The design of the policy network results in policy parameter sharing among AVs, which drastically reduces the exploration space during training. Moreover, it vectorizes the inference for all the AVs in one scenario.

    The critic network uses the same embedded features as query, key, and value for the attention layers. The network mainly consists of a stack of two identical attention layers. Each layer has a self-attention module and a feed-forward network. The residual connections and layer normalization are the same as those in the policy network. For the self-attention calculation, $M_s$ is utilized to guarantee that the vehicles only attend to existing ones.
    The encoded features of all vehicles are reduced to a single vector with the pooling operation $\sum$, which accommodates different numbers of vehicles.
    After the last linear operation, it outputs the state value $V(s_t)$.

\section{Experiments}
\label{sec:experiments}
In this section, we present the training and evaluation results with different maps and penetration rates of AVs.
Afterwards, we explore the issue of conservative AVs in particular with respect to the decentralized policies.

\subsection{Experiment Setup}
    The microscopic traffic simulator SUMO~\cite{SUMO2018} is utilized for training and evaluation. The most typical bottleneck locations in road systems shown in Fig.~\ref{fig:maps} are created to test the utility of the proposed method:
    \begin{itemize}[leftmargin=*]
        \item[] \textbf{On-ramp}: Vehicles from the on-ramp merge into a two-lane freeway. Freeway vehicles have the right of way.
        \item[] \textbf{Four-way intersection}: The main road involving the upper and the bottom edges has higher priority over the side road.
        \item[] \textbf{Three-way intersection}: The main road, which consists of the left and the right edges, has higher priority.
        \item[] \textbf{Lane drop}: This map is the same as the most relevant work~\cite{vinitsky2023optimizing}. Four lanes converge into two lanes, which then merge into a single lane. The priority of all the lanes is equal so that vehicles normally take turns merging into the new lane in ``zipper'' fashion.
    \end{itemize}

    \subsubsection{Traffic Episodes for Training and Evaluation}
    To make it easier to reproduce the results and compare different methods, we generate two collections of traffic episodes for training and evaluation.
    The approach to generating these episodes is similar to that of our previous work~\cite{yan2021courteous}. Each of them has a duration of $\qty{1200}{\second}$. We sample the input traffic flow and its distribution on different routes randomly. It can vary drastically during the whole episode. Vehicles can turn into any connected lanes with a randomly sampled turning rate. All these methods result in complex and relatively realistic traffic conditions, exposing great challenges for the learning algorithm.
    The training traffic data for each map comprises $\num{5000}$ episodes, while the evaluation data for each traffic condition includes $\num{20}$ episodes.

    \subsubsection{Baselines}
    Three traffic controllers are employed to compare with our proposed decentralized controller, which we refer to as decentralized vehicle controller (\textbf{DVC}) for simplicity in data visualization. 1) No controller (\textbf{NC}): all the vehicles only follow the driver models in SUMO. 2) Intelligent traffic signal control~(\textbf{ITSC}): courteous virtual traffic signal control (\textbf{CVTSC})~\cite{yan2021courteous} with full observability is implemented for intersections and the results of feedback controller \textbf{ALINEA}~\cite{vinitsky2023optimizing} is used for lane drop. 3) Centralized vehicle controller (\textbf{CVC}): based on the models shown in Fig.~\ref{fig:policy} and Fig.~\ref{fig:critic} we design a centralized controller, where every AV has full observability. The critic network is the same as the decentralized controller. In the policy network, instead of keeping the key and value unchanged in the attention layers, the features of the activated AVs are updated along with the query values.
    As a result, the activated AVs can not only observe all the other vehicles but also communicate implicitly with each other in the policy.

    \subsubsection{Implementation}
    We employ the PPO algorithm from RLLib 2.4.0~\cite{liang2018rllib} with 32 workers for policy rollout. Training converges within 2 to 4 hours on a desktop equipped with 24 CPU cores and an NVIDIA RTX 4060 Ti GPU.
    Each setting is trained with five randomly chosen seeds $[8715736, 21320071, 27631279, 38961730, 88104531]$.
    The discount factor $\gamma=0.98$.
    Other hyper-parameters, including equity factors, batch size, mini-batch size, value loss coefficient, entropy coefficient, and learning rate, are selected through grid search for CVC with an AV penetration rate of $30\%$.
    The proposed method converges well in all settings across different seeds.
    Our code is publicly available at \url{https://github.com/shengchao-y/MAAAC-driving}.

\subsection{Mitigating Congestion with Autonomous Vehicles}
    \begin{table*}[t]
    \vspace{5pt}
    \centering
    \begin{tabular}{lllllllllllllllll}
        \toprule
         \multirow{2}{*}{\shortstack{Controller}} & \multicolumn{4}{c}{3-way intersection} & \multicolumn{4}{c}{on-ramp} & \multicolumn{4}{c}{4-way intersection} & \multicolumn{4}{c}{lane drop} \\
         & 1000 & 1500 & 2000 & 2500 & 3000 & 3500 & 4000 & 4500 & 1000 & 1500 & 2000 & 2500 & 1500 & 2000 & 2500 & 3000 \\
        \midrule
        NC & 98.2 & 84.9 & 65.5 & 53.6 & 97.9 & 96.5 & 88.3 & 79.0 & 98.3 & 85.0 & 67.2 & 54.0 & 99.4 & 78.1 & 62.3 & 51.9 \\
        \cmidrule(lr){2-5} \cmidrule(lr){6-9} \cmidrule(lr){10-13} \cmidrule(lr){14-17}
        DVC-5-100 & 98.2 & 90.3 & 74.8 & 61.9 & 97.9 & 96.6 & 88.5 & 79.2 & 98.1 & 87.8 & 70.2 & 57.5 & 98.9 & 92.9 & 75.6 & 63.0 \\
        DVC-10-100 & 98.5 & 93.3 & 78.4 & 65.0 & 97.9 & 96.7 & 88.8 & 79.3 & 98.0 & \textbf{91.1} & 72.5 & 58.8 & 99.5 & 95.1 & 78.0 & 62.3 \\
        DVC-20-100 & 98.4 & 94.6 & 78.9 & 65.2 & 97.9 & 96.8 & 89.0 & 79.5 & 97.9 & 90.9 & 72.1 & 59.1 & 99.6 & 97.6 & 79.2 & 66.7 \\
        DVC-40-100 & 98.2 & 92.9 & 77.9 & 68.1 & 97.9 & 96.9 & 89.5 & 79.9 & 98.2 & 91.0 & \textbf{72.6} & 58.4 & 99.1 & \cellcolor{ForestGreen!25}{99.9} & 81.8 & 67.6 \\
        DVC-80-100 & 98.5 & \textbf{95.0} & \textbf{81.2} & 67.8 & 97.9 & \cellcolor{ForestGreen!25}{97.1} & \cellcolor{ForestGreen!25}{90.5} & \cellcolor{ForestGreen!25}{80.8} & 98.0 & 89.9 & 71.4 & 58.8 & 99.5 & \cellcolor{ForestGreen!25}{99.9} & \cellcolor{ForestGreen!25}{84.5} & 67.0 \\
        \cmidrule(lr){2-5} \cmidrule(lr){6-9} \cmidrule(lr){10-13} \cmidrule(lr){14-17}
        CVC-5 & 98.2 & 92.6 & 77.8 & 64.0 & 97.8 & 96.7 & 88.6 & 79.2 & 97.2 & 89.1 & 70.3 & 58.4 & 99.3 & 94.1 & 76.2 & 62.0 \\
        CVC-10 & 98.4 & 93.3 & 79.2 & 66.5 & 97.9 & 96.8 & 88.8 & 79.4 & 98.0 & 90.0 & 71.4 & 58.8 & 99.3 & 96.2 & 78.0 & 64.5 \\
        CVC-20 & 98.4 & 95.3 & 80.6 & 67.6 & 97.8 & 96.8 & 89.2 & 79.8 & 97.5 & 89.0 & 69.6 & 58.7 & 99.1 & 98.9 & 81.2 & 65.7 \\
        CVC-40 & 97.6 & 86.2 & 72.9 & 60.6 & 97.9 & \textbf{97.0} & \textbf{89.7} & 80.0 & 98.0 & 90.4 & 71.0 & 59.1 & 99.1 & \textbf{99.0} & 81.2 & \textbf{67.8} \\
        CVC-80 & 98.2 & 92.8 & 77.4 & 63.9 & 97.9 & \cellcolor{ForestGreen!25}{97.1} & \cellcolor{ForestGreen!25}{90.5} & \cellcolor{ForestGreen!25}{80.8} & 97.9 & 88.6 & 70.5 & 57.8 & 99.6 & \textbf{99.0} & \textbf{83.4} & 61.6 \\
        \cmidrule(lr){2-5} \cmidrule(lr){6-9} \cmidrule(lr){10-13} \cmidrule(lr){14-17}
        DVC-5-50 & 98.2 & 92.1 & 76.2 & 63.8 & 97.9 & 96.6 & 88.5 & 79.0 & 98.2 & 88.1 & 68.0 & 57.5 & 99.6 & 93.2 & 75.0 & 60.8 \\
        DVC-10-50 & 98.5 & 92.5 & 76.4 & 64.4 & 97.8 & 96.7 & 88.6 & 79.2 & 98.3 & 90.5 & 72.5 & \textbf{59.6} & 99.9 & 85.3 & 70.6 & 60.6 \\
        DVC-20-50 & 98.3 & 93.3 & 79.3 & 67.1 & 97.9 & 96.8 & 88.8 & 79.5 & 98.4 & 89.7 & 72.5 & 59.1 & 99.7 & 84.7 & 69.2 & 61.9 \\
        DVC-40-50 & 98.4 & 93.7 & 80.1 & \textbf{69.1} & 97.9 & \textbf{97.0} & \textbf{89.7} & \textbf{80.1} & 97.7 & 84.8 & 69.9 & 57.4 & 99.9 & 86.7 & 71.8 & 60.3 \\
        DVC-80-50 & 98.4 & 89.7 & 74.4 & 64.6 & 97.9 & \cellcolor{ForestGreen!25}{97.1} & \cellcolor{ForestGreen!25}{90.5} & \cellcolor{ForestGreen!25}{80.8} & 98.3 & 89.1 & 71.2 & 58.8 & 99.4 & 82.0 & 71.4 & 59.0 \\
        \cmidrule(lr){2-5} \cmidrule(lr){6-9} \cmidrule(lr){10-13} \cmidrule(lr){14-17}
        ITSC & 98.4 & \cellcolor{ForestGreen!25}{97.2} & \cellcolor{ForestGreen!25}{92.7} & \cellcolor{ForestGreen!25}{86.1} & - & - & - & - & 98.2 & \cellcolor{ForestGreen!25}{93.7} & \cellcolor{ForestGreen!25}{84.9} & \cellcolor{ForestGreen!25}{70.8} & 99.6 & 97.8 & 82.4 & \cellcolor{ForestGreen!25}{68.7} \\
        \bottomrule
    \end{tabular}
    \caption{
    Throughput ($\%$) comparison across different maps and traffic inputs under various control schemes. Throughput, defined as the ratio of output to input traffic flow, is computed as an average across $20$ episodes and five seeds. Traffic input is measured in vehicles per hour. Policy names, denoting the trained controllers, concatenate the AV penetration rate and the observation range for clarity. For instance, \textit{DVC-5-100} indicates a policy with a $5\%$ AV penetration rate and a $\qty{100}{\meter}$ observation distance. \textit{ITSC} refers to intelligent traffic signal control, employing CVTSC~\cite{yan2021courteous} with an $80\%$ AV penetration at intersections and ALINEA~\cite{vinitsky2023optimizing} for lane drops. Within congested scenarios, the \hl{highest} throughput (column-wise) is marked in green background, while the \textbf{second-highest} is in bold.
    }
    \label{tab:tp}
    \vspace{-5pt}
\end{table*}

    According to previous research, the performance of decentralized traffic management controllers depends on the penetration rate and the observation range.
    To evaluate their effect on the proposed methods, we train and evaluate the controllers for each map with $5$ different penetration rates of AV in $\{5\%,10\%,20\%,40\%,80\%\}$. The AV observation range is $\qty{100}{\meter}$ for these environments. Additionally, we train the decentralized controllers with all five penetration rates and an observation range of $\qty{50}{\meter}$.
    The varying observation range does not impact the centralized controllers due to their full observability.
    The results comparing the performance of different methods in different environments are shown in Table~\ref{tab:tp}. 
    Both CVC and DVC can improve the throughput in all scenarios compared with the simple baseline with no high-level controller.
    Besides, several interesting results can be observed in the data.

    \subsubsection{Comparison to MARL Results}
    In previous studies on lane drop scenarios, the traffic signal controller ALINEA
    was found to be the best-performing approach~\cite{vinitsky2023optimizing}. Our decentralized policies, developed through a centralized training algorithm, not only outperform those MARL methods but also exceed the performance of ALINEA under certain traffic conditions. These findings validate our hypothesis that algorithms designed to adapt to fluctuating traffic inputs can generate more effective policies compared to MARL strategies that assume static traffic flow conditions.

    \subsubsection{Performance Degradation at Higher Penetration Rates}
    Unlike the outcomes observed in our prior work on a virtual traffic signal controller~\cite{yan2021courteous}, where throughput monotonically increased with the autonomous vehicle (AV) penetration rate, we observe a performance degradation under certain conditions with the proposed vehicle controllers. This phenomenon is most prominent at a high penetration rate of $80\%$. Directly controlling individual AVs results in a substantially larger action space for the entire system compared to traffic signal control, which manages the intersection with a small set of signals. In scenarios such as lane drops, where up to $16$ AVs may be active, this corresponds to a joint action space size of $6^{16}\approx \num{2.8e+12}$. Although parameter sharing can help mitigate the expansion of the exploration space, it nonetheless grows exponentially with the number of activated AVs. We hypothesize that the expansion of the search space introduces substantial challenges to the training process.
    As a result, ITSC achieves the best performance for intersections.

    \subsubsection{Observation Range Impact on Performance}
    The impact of the observation range, specifically $\qty{100}{\meter}$ versus $\qty{50}{\meter}$, on performance is subtle, except for the lane drop scenario. In lane drops, where higher speed limits result in increased vehicle separation, the benefit of a larger observation distance is higher.
    Conversely, in scenarios characterized by closer vehicle proximity, the additional data from an extended observation range does not yield a clear advantage and may detract from overall performance.

    \subsubsection{Centralized Control Not Always Superior}
    Although CVC agents leverage global state information to control individual vehicles, this approach does not inherently yield better results than decentralized methods and can often result in poorer performance. The benefits of centralized policies are largely confined to specific scenarios, like lane drops, where an extended observation range can substantially enhance performance.
    However, in situations in which an increased observation range does not offer a clear advantage, centralized controllers perform comparably or are even less effective, indicating that enhanced information exchange among AVs does not necessarily contribute to improved policy efficacy.
    We assume the performance drop of CVC agents is caused by the increased observation space introduced by the further vehicles.
    They make the optimization more complex while bringing little useful information.

    \begin{table}[t]
    \vspace{5pt}
    \centering
    \begin{tabular}{lllll}
    \toprule
    \multirow{2}{*}{\shortstack{Controller}} & \multicolumn{4}{c}{on-ramp} \\
     & 3000 & 3500 & 4000 & 4500 \\
    \midrule
    NC & 14.4 & 37.0 & 173.7 & 326.1 \\
    \cmidrule(lr){2-5}
    DVC-5-100 & 14.0 & 34.9 & 169.0 & 321.1 \\
    DVC-10-100 & 13.9 & 33.1 & 166.3 & 320.7 \\
    DVC-20-100 & 13.8 & 32.4 & 164.2 & 316.0 \\
    DVC-40-100 & 13.9 & 31.6 & 153.7 & 309.2 \\
    DVC-80-100 & 14.2 & \textbf{29.3} & \cellcolor{ForestGreen!25}{139.0} & \cellcolor{ForestGreen!25}{294.9} \\
    \cmidrule(lr){2-5}
    CVC-5 & 14.1 & 35.9 & 169.6 & 322.7 \\
    CVC-10 & 13.9 & 30.9 & 164.2 & 316.8 \\
    CVC-20 & 14.2 & 31.7 & 157.8 & 311.4 \\
    CVC-40 & 14.3 & 29.3 & 152.4 & 309.3 \\
    CVC-80 & 14.2 & \textbf{29.3} & \cellcolor{ForestGreen!25}{139.0} & \cellcolor{ForestGreen!25}{294.9} \\
    \cmidrule(lr){2-5}
    DVC-5-50 & 14.0 & 34.6 & 171.9 & 325.8 \\
    DVC-10-50 & 14.4 & 35.9 & 171.3 & 322.7 \\
    DVC-20-50 & 14.0 & 33.2 & 165.6 & 316.7 \\
    DVC-40-50 & 13.9 & \cellcolor{ForestGreen!25}{27.7} & \textbf{150.3} & \textbf{307.5} \\
    DVC-80-50 & 14.2 & \textbf{29.3} & \cellcolor{ForestGreen!25}{139.0} & \cellcolor{ForestGreen!25}{294.9} \\
    \bottomrule
    \end{tabular}
    \caption{Average waiting time $T_\mathrm{wait}$ in seconds of unreleased vehicles at on-ramp under various traffic inputs and controllers, corresponding the evaluation in Table~\ref{tab:tp}. In congested scenarios, the \hl{best} results are marked with green background, and the \textbf{second-best} outcomes are in bold.
    }
    \label{tab:t_wait}
    \vspace{-5pt}
\end{table}
    \pgfplotstableread{
x       y       y-min   y-max
NC-5      22.67   1.50    1.41
NC-10      22.60   1.39    1.52    
NC-20      22.48   1.34    1.48
NC-40      22.10   1.30    1.47
NC-80      21.75   1.23    1.32
}{\avmainttNC}

\pgfplotstableread{
x       y       y-min   y-max
NC-5      213.91  128.53  186.49
NC-10      148.48  98.69   225.38
NC-20      237.87  119.51  178.85
NC-40      376.25  161.73  164.91
NC-80      422.42  190.06  239.27  
}{\avrampttNC}

\pgfplotstableread{
x       y       y-min   y-max
NC-0      21.97   1.25    1.40
NC-5      22.06   1.28    1.42
NC-10      22.12   1.29    1.42
NC-20      22.08   1.25    1.42
NC-40      21.94   1.22    1.37
NC-80      21.78   1.22    1.36
}{\hvmainttNC}

\pgfplotstableread{
x       y       y-min   y-max
NC-0      25.31   2.04    4.96
NC-5      26.22   2.62    7.88
NC-10      29.02   4.69    14.68
NC-20      48.49   21.50   137.21
NC-40      156.93  125.88  202.22
NC-80      255.54  233.07  251.44
}{\hvrampttNC}

\pgfplotstableread{
x       y
NC-5      98.22   
NC-10      98.18
NC-20      98.20
NC-40      98.17   
NC-80      98.28
}{\avmaintpNC}

\pgfplotstableread{
x       y
NC-5      42.03
NC-10      44.43
NC-20      45.04
NC-40      36.69
NC-80      22.61
}{\avramptpNC}

\pgfplotstableread{
x       y
NC-0      98.24
NC-5      98.24
NC-10      98.23
NC-20      98.25
NC-40      98.24
NC-80      98.20
}{\hvmaintpNC}

\pgfplotstableread{
x       y
NC-0      97.08
NC-5      96.73
NC-10      90.24
NC-20      71.91
NC-40      47.76
NC-80      32.86
}{\hvramptpNC}
\pgfplotstableread{
x       y       y-min   y-max
DVC-5      22.86   1.69   2.31
DVC-10     25.50   3.33    2.97
DVC-20      24.17   2.11    2.31
DVC-40      23.58   1.81    2.02
DVC-80      23.22   1.70    1.93
}{\avmaintt}

\pgfplotstableread{
x       y       y-min   y-max
DVC-5      148.72   81.02   132.28
DVC-10      57.32   19.04   37.71
DVC-20      48.19   14.30     32.83
DVC-40      43.77   11.37   23.49
DVC-80      58.18   20.09   34.14
}{\avramptt}

\pgfplotstableread{
x       y       y-min   y-max
DVC-5      22.02    1.26    1.43
DVC-10      22.26   1.35    1.53
DVC-20      22.30   1.34    1.54
DVC-40      22.38   1.35    1.49
DVC-80      22.46   1.45    1.46
}{\hvmaintt}

\pgfplotstableread{
x       y       y-min   y-max
DVC-5      25.84   2.43   6.54
DVC-10      26.45   2.68   5.34
DVC-20      26.77   2.80    5.66
DVC-40      27.20   2.91    5.60
DVC-80      32.02   5.88    15.90
}{\hvramptt}

\pgfplotstableread{
x       y
DVC-5    98.2
DVC-10    98.0
DVC-20    98.0
DVC-40    98.1
DVC-80    98.19
}{\avmaintp}

\pgfplotstableread{
x       y
DVC-5    84.1
DVC-10    91.8
DVC-20    94.0
DVC-40    94.0
DVC-80    90.31
}{\avramptp}

\pgfplotstableread{
x       y
DVC-5    98.2
DVC-10    98.2
DVC-20    98.2
DVC-40    98.3
DVC-80    98.13
}{\hvmaintp}

\pgfplotstableread{
x       y
DVC-5    96.9
DVC-10    97.0
DVC-20    96.9
DVC-40    96.9
DVC-80    94.81
}{\hvramptp}

\pgfplotstableread{
x       y
NC-0      2000
}{\plotlabel}

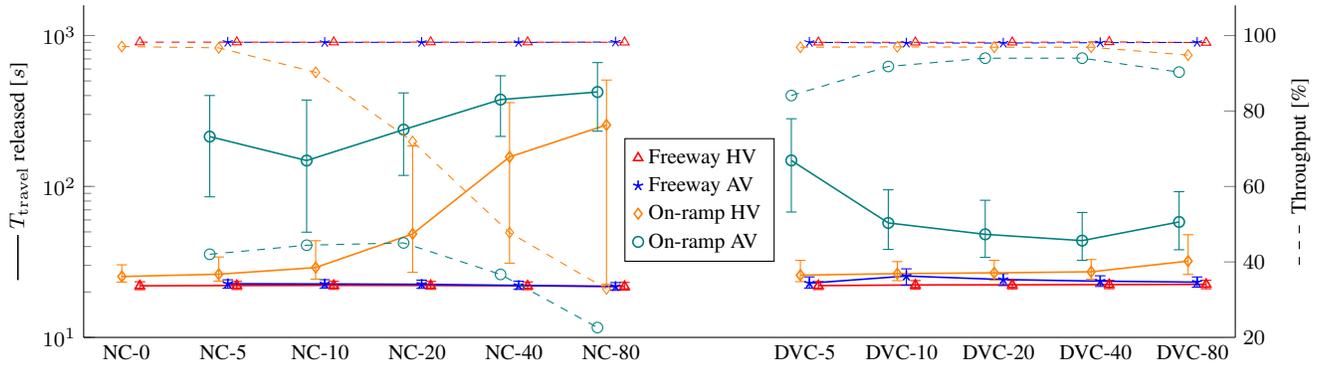
\begin{figure*}[t]
	\centering
	\scalefont{0.8}
	\begin{tikzpicture}
	\begin{axis} [axis on top,
			height=6cm,
			ymode=log,
			width=0.95\textwidth,
			bar width=0.15cm,
			enlarge y limits={value=.1,upper},
			ymin=1e1, ymax=1e3,
			axis x line*=bottom,
			axis y line*=left,
			xtick style={draw=none},
			enlarge x limits={value=0.04, auto}, 
			ylabel={\ref{plt:ttplot} $T_\mathrm{travel}$ released [$s$]},
			symbolic x coords={NC-0, NC-5, NC-10, NC-20, NC-40, NC-80, , DVC-5, DVC-10, DVC-20, DVC-40, DVC-80},
                xtick={NC-0, NC-5, NC-10, NC-20, NC-40, NC-80,, DVC-5, DVC-10, DVC-20, DVC-40, DVC-80},
                legend style={only marks, at={(0.6,0.6)},anchor=north east}, legend cell align={left},legend entries={Freeway HV, Freeway AV, On-ramp HV, On-ramp AV}
                ]
            \addplot[ mark=triangle,semithick,mark options={scale=1}, xshift=0.18cm,legend image post style={xshift=-0.18cm}, color=red, error bars/.cd, y dir=both, y explicit]
            table[x=x,y=y,y error plus expr=\thisrow{y-max},y error minus expr=\thisrow{y-min}] {\hvmainttNC};
            \addplot[ mark=star,semithick,mark options={scale=1}, xshift=0.06cm,legend image post style={xshift=-0.06cm}, color=blue, error bars/.cd, y dir=both, y explicit]
            table[x=x,y=y,y error plus expr=\thisrow{y-max},y error minus expr=\thisrow{y-min}] {\avmainttNC};
            \addplot[ mark=diamond,semithick,mark options={scale=1}, xshift=-0.06cm,legend image post style={xshift=0.06cm}, color=orange, error bars/.cd, y dir=both, y explicit]
            table[x=x,y=y,y error plus expr=\thisrow{y-max},y error minus expr=\thisrow{y-min}] {\hvrampttNC};
            \addplot[ mark=o,semithick,mark options={scale=1}, xshift=-0.18cm,legend image post style={xshift=0.18cm}, color=teal, error bars/.cd, y dir=both, y explicit]
            table[x=x,y=y,y error plus expr=\thisrow{y-max},y error minus expr=\thisrow{y-min}] {\avrampttNC};
            \addplot[ mark=none,semithick, color=black]
            table[x=x,y=y] {\plotlabel};\label{plt:ttplot}
            \addplot[ mark=triangle,semithick,mark options={scale=1}, xshift=0.18cm, color=red, error bars/.cd, y dir=both, y explicit]
            table[x=x,y=y,y error plus expr=\thisrow{y-max},y error minus expr=\thisrow{y-min}] {\hvmaintt};
            \addplot[ mark=star,semithick,mark options={scale=1}, xshift=0.06cm, color=blue, error bars/.cd, y dir=both, y explicit]
            table[x=x,y=y,y error plus expr=\thisrow{y-max},y error minus expr=\thisrow{y-min}] {\avmaintt};
            \addplot[ mark=diamond,semithick,mark options={scale=1}, xshift=-0.06cm, color=orange, error bars/.cd, y dir=both, y explicit]
            table[x=x,y=y,y error plus expr=\thisrow{y-max},y error minus expr=\thisrow{y-min}] {\hvramptt};
            \addplot[ mark=o,semithick,mark options={scale=1}, xshift=-0.18cm, color=teal, error bars/.cd, y dir=both, y explicit]
            table[x=x,y=y,y error plus expr=\thisrow{y-max},y error minus expr=\thisrow{y-min}] {\avramptt};
	\end{axis} 
	
	\begin{axis}[axis on top,
			height=6cm, width=0.95\textwidth,
			bar width=0.15cm,
			enlarge y limits={value=.1,upper},
			ymin=20, ymax=100,
			axis x line*=bottom,
			axis y line*=right,
			xtick style={draw=none},
			subtickwidth=0pt,
			enlarge x limits={value=0.04, auto}, 
			ylabel={\ref{plt:tpplot} Throughput [\%]},
			symbolic x coords={NC-0, NC-5, NC-10, NC-20, NC-40, NC-80,, DVC-5, DVC-10, DVC-20, DVC-40, DVC-80},
			xtick=\empty,
			x axis line style={draw=none}]
	\addplot[ mark=triangle,mark options={scale=1,solid}, xshift=0.18cm, color=red, dashed]
	table[x=x,y=y] {\hvmaintpNC};
	\addplot[ mark=star,mark options={scale=1,solid}, xshift=0.06cm, color=blue, dashed]
	table[x=x,y=y] {\avmaintpNC};
	\addplot[ mark=diamond,mark options={scale=1,solid}, xshift=-0.06cm, color=orange, dashed]
	table[x=x,y=y] {\hvramptpNC};
        \addplot[ mark=o,mark options={scale=1,solid}, xshift=-0.18cm, color=teal, dashed]
	table[x=x,y=y] {\avramptpNC};
	\addplot[ mark=none, color=black, dashed]
	table[x=x,y=y] {\plotlabel};\label{plt:tpplot}
        \addplot[ mark=triangle,mark options={scale=1,solid}, xshift=0.18cm, color=red, dashed]
	table[x=x,y=y] {\hvmaintp};
	\addplot[ mark=star,mark options={scale=1,solid}, xshift=0.06cm, color=blue, dashed]
	table[x=x,y=y] {\avmaintp};
	\addplot[ mark=diamond,mark options={scale=1,solid}, xshift=-0.06cm, color=orange, dashed]
	table[x=x,y=y] {\hvramptp};
        \addplot[ mark=o,mark options={scale=1,solid}, xshift=-0.18cm, color=teal, dashed]
	table[x=x,y=y] {\avramptp};
	\end{axis}
	
    \end{tikzpicture}
    \caption{
    Comparative analysis of traffic flow for different vehicle groups on the on-ramp map with a traffic input of $\num{3500}\nicefrac{\text{v}}{\si{\hour}}$.
    Travel times ($T_\mathrm{travel}$) represent the median, lower, and upper quartiles for all vehicles successfully exiting the system across $20$ evaluative episodes.
    Throughput quantifies the proportion of vehicles exiting versus the total vehicles introduced during these episodes.
    "NC-$x$" denotes scenarios without a controller at $x\%$ AV penetration, while "DVC-$x$" refers to scenarios employing our developed decentralized policy at the corresponding penetration rate.
    }
    \label{fig:groups_together}
    \vspace{-5pt}
\end{figure*}

    \subsubsection{On-Ramp Dynamics}
    In the on-ramp scenario, there are only minor throughput improvements. This could be attributed to the parallel merging lanes
    offering vehicles more cooperation flexibility compared to other scenarios where cooperative maneuvers are confined to narrow spaces at junctions or lane drop endpoints. Consequently, under the default driver model settings in SUMO, the scope for augmenting throughput on on-ramps appears constrained. However, the mitigation of congestion is evident, which can be demonstrated by the reduced average waiting times for vehicles yet to be released (see  Table~\ref{tab:t_wait}).  At the end of an episode, the waiting time $t_\mathrm{wait}$ for any vehicle not having traversed the map is computed as the episode duration minus the scheduled entry time for that vehicle. The obvious reduction in average waiting times underlines the efficacy of our approach in alleviating congestion.

\subsection{Too Cautious to Drive?}
    
    In SUMO, the driver model is defined with various parameters that influence vehicle behavior in car-following and lane-changing situations. Parameters such as $lcAssertive$ and $lcSpeedGain$ indicate the driver's aggressiveness level. Specifically, $lcAssertive$ quantifies a driver's tendency to accept smaller front and rear gaps on the target lane during a lane change, whereas $lcSpeedGain$ reflects the driver's inclination to change lanes for potential speed benefits. Vehicles characterized by lower values of these parameters are deemed more conservative compared to the default settings of $lcAssertive=1$ and $lcSpeedGain=1$.
    To explore the impact of conservative autonomous vehicles (AVs) on traffic, we conducted training sessions in environments with $lcAssertive=0.1$ and $lcSpeedGain=0$ for AVs. Furthermore, to mirror the design emphasis on comfort and smooth driving experiences typically associated with AVs, we reduced their maximum deceleration and acceleration from $2.6$ and $\qty{4.5}{\meter\per\square\second}$ to $2$ and $\qty{3.5}{\meter\per\square\second}$, respectively. This analysis focuses on the on-ramp scenario.

    Fig.~\ref{fig:groups_together} illustrates the throughput and travel time of the released vehicles categorized in four groups under traffic input $\num{3500}\nicefrac{\text{v}}{\si{\hour}}$. With no controller~(NC), we note a substantial increase in congestion within the on-ramp lane as the presence of AVs in the traffic rises. At a mere $5\%$ AV penetration rate, AVs on the on-ramp lane begin queuing, awaiting their turn to merge into the freeway. Meanwhile, HVs in the on-ramp lane manage to change lanes by forcing freeway vehicles to slow down, effectively bypassing the queued AVs. However, as AV penetration increases, the queue lengthens, eventually obstructing the entire merging lane and preventing HVs from accessing the merging zone. Consequently, the throughput for vehicles on the on-ramp lane reduces to approximately $30\%$ with an $80\%$ AV penetration rate, accompanied by a substantial increase in their travel time.
    
    This analysis underlines public apprehensions regarding AVs that, despite being engineered for safety and efficiency, may inadvertently impair traffic flow. To anticipate these issues, some researchers advocate for developing policies that trade off safety for efficiency, emulating the more assertive driving styles of human drivers. However, replicating human-like driving behavior presents not only technical hurdles, such as forecasting the movements of other vehicles but also ethical concerns. Specifically, assigning responsibility becomes complex when an algorithm deliberately prioritizes speed over safety, potentially resulting in accidents.

    Fortunately, we offer a promising alternative to address this issue. The plots on the right side of Fig.~\ref{fig:groups_together} illustrate that our cooperative driving strategies can mitigate congestion caused by cautious AVs. Although these vehicles experience longer travel times than their more assertive counterparts, cooperation among AVs significantly narrows this gap. We observe a slight increase in travel time for AVs on the freeway, which helps keep the merging zone clear and provides vehicles more opportunities to merge into traffic. Remarkably, the throughput for vehicles on the on-ramp lane increases threefold with an $80\%$ AV penetration rate. This substantial outcome indicates a feasible future where conservative, cooperative AVs can deliver both safety and efficiency.

\section{Conclusion and Limitations}

In this paper, we consider the problem of improving traffic flow at bottlenecks of the road system through a decentralized control approach for automated vehicles with partial observability. 
To solve this problem, we introduce an asymmetric actor-critic model structure, trained using single-agent reinforcement learning. 
By treating each AV as a token of the state for the entire system rather than as a distinct agent, this method can learn decentralized policies for individual AVs in infinite-horizon traffic with dynamic flow input. The evaluation against baseline controllers across different bottleneck locations shows that our model substantially improves the traffic flow.
The experiments further demonstrate that cooperative autonomous vehicles can mitigate the problem of reduced traffic flow caused by their strict adherence to traffic rules. 

Despite the advancements, several aspects warrant future research.
For example, the exploration problem induced by the extensive action space of controlling individual AVs presents a challenge, especially at high AV penetration rates. Future research could focus on developing methods to selectively activate a smaller number of AVs for control. 
This adjustment could potentially enhance policy effectiveness by focusing on AVs that substantially influence traffic flow, considering that many vehicles merely follow their leaders in traffic.

\addtolength{\textheight}{-7cm}   

\bibliographystyle{IEEEtran}
\bibliography{mavic}

\end{document}